%% file: main.tex
\documentclass[11pt]{article}
\usepackage{fullpage}

\usepackage[utf8]{inputenc} 
\usepackage{hyperref}       
\usepackage{url}            
\usepackage{booktabs}       
\usepackage{amsfonts}       
\usepackage{microtype}      
\usepackage{xcolor}         
\usepackage{graphicx}
\usepackage{float}
\usepackage{amsmath, amsthm,amsfonts}
\usepackage[square, numbers]{natbib}
\usepackage{bm, bbm}

\newtheorem{theorem}{Theorem}

\newtheorem{lemma}{Lemma}

\newtheorem{fact}{Fact}

\theoremstyle{remark}

\input{notations}

\hypersetup{
    colorlinks = true,
 	allcolors = teal,
}

\title{How Does Independence Help Generalization? \\ Sample Complexity of ERM on Product Distributions}

\author{
  Tao Lin \\
  Harvard University \\ 
  \texttt{tlin@g.harvard.edu}
}

\begin{document}

\maketitle


\begin{abstract}
While many classical notions of learnability (e.g., PAC learnability) are distribution-free, utilizing the specific structures of an input distribution may improve learning performance.
For example, a product distribution on a multi-dimensional input space has a much simpler structure than a correlated distribution. 
A recent paper \cite{guo2021generalizing} shows that the sample complexity of a general learning problem on product distributions is polynomial in the input dimension, which is exponentially smaller than that on correlated distributions.  However, the learning algorithm they use is not the standard Empirical Risk Minimization (ERM) algorithm.   
In this note, we characterize the sample complexity of ERM in a general learning problem on product distributions. 
We show that, even though product distributions are simpler than correlated distributions, ERM still needs an exponential number of samples to learn on product distributions, instead of a polynomial.
This leads to the conclusion that a product distribution by itself does not make a learning problem easier -- an algorithm designed specifically for product distributions is needed.

\end{abstract}

\input{body.tex}
\bibliographystyle{alpha}
\bibliography{bibfile}

\appendix
\input{appendix.tex}

\end{document}

%% file: notations.tex
\DeclareMathOperator*{\argmin}{arg\,min}

\newcommand{\E}{{\mathbb E}}
\newcommand{\Var}{\mathrm{Var}}

\newcommand{\eps}{\varepsilon}

\newcommand{\ERM}{\mathrm{ERM}}
\newcommand{\emp}{\mathrm{emp}}
\newcommand{\dTV}{d_{\mathrm{TV}}}

%% file: body.tex
\section{Introduction}
While many classical notions of learnability (e.g., PAC learnability, VC dimension) are distribution-free, utilizing the specific structure of a distribution may improve learning performance.
A recent paper \cite{guo2021generalizing} suggests that, the sample complexity (or equivalently, generalization error) of some learning problems on a multi-dimensional input space can be significantly reduced if the input distribution is a \emph{product} distribution, instead of an arbitrarily correlated distribution.
Building on that, this note aims to explore the difference between correlated distributions and product distributions as the input distribution to a general learning problem.

\paragraph{Examples of learning problems with product distributions} 
Although product distributions do not fit into the standard supervised learning framework where a sample $s=(x, y)$ from the underlying distribution $D$ consists of a feature part $x\in\mathcal X$ and a correlated label part $y\in\mathcal Y$, there are many unsupervised learning problems with product distributions. 
One such example, which is also an application of statistical learning theory to economics, is \emph{auction design} \citep{cole_sample_2014}.  A seller auctions off an item to several bidders each of whom has a valuation $v^j$ for the item drawn from some unknown distribution $D^j$ ($j=1, \ldots, n$, where $n\ge 2$ is the number of bidders). 
It is standard in the auction design literature to assume that bidders have independent valuations, so the distributions $D^1, \ldots, D^n$ are not correlated. 
The seller have some samples of bidders' valuations, based on which she wants to design an auction to maximize her expected revenue with respect to the unknown valuation distribution $D = D^1 \times \cdots \times D^n$.
This is an unsupervised learning problem with a product input distribution $D$, where each sample $s_i$ from $D$ consists of the independent valuations of all bidders: $s_i=(v_i^1, \ldots, v_i^n)$.  Other examples of learning problems with product distributions include, e.g., prophet inequalities \cite{correa_prophet_2019} and the Pandora's box problem \cite{guo2021generalizing, fu_learning_2020}.    

\paragraph{Contribution of this note}
This note explores the role of product distributions in reducing the sample complexity of a general unsupervised learning problem.
As in \cite{guo2021generalizing}, we take the \emph{distribution learning} problem as the general learning problem: given samples from an underlying distribution $D$, we want to construct a distribution $E$ that approximates $D$.

The standard way to estimate a distribution from samples is to use the \emph{empirical distribution}, which corresponds to the canonical Empirical Risk Minimization (ERM) principle in learning theory. In fact, using the empirical distribution \emph{is} the optimal way to estimate a distribution $D$ about which we do not have any prior knowledge \citep{han2015minimax}.
However, when $D$ is an arbitrarily correlated distribution on an $n$-dimensional space, it may take an \emph{exponential} (in $n$) number of samples for the empirical distribution to be a good approximation to $D$ (see Section~\ref{sec:correlated_exp} for details). 

When $D$ is a product distribution, on the other hand, \cite{guo2021generalizing} show that the number of samples needed to obtain a good approximation to $D$ can be significantly reduced, in particular, to a \emph{polynomial} in $n$. 
However, the way they estimate $D$ is not the standard way of using the empirical distribution.  Instead, they first estimate each independent component $D^j$ of $D$ by the empirical distribution $E^j$ on the samples from $D^j$ and then use the product $E = E^1 \times \cdots \times E^n$ as the estimate of $D = D^1\times \cdots \times D^n$.  They call such an estimate $E$ a \emph{product empirical distribution}. (See Section~\ref{sec:product_PERM_poly} for details.)  

In this note, we ask the following questions: \emph{What} enables the exponential reduction of sample complexity when the distribution $D$ switches from an arbitrarily correlated distribution to a product distribution?  Is it only because product distributions have simpler structures?  Or is it also related to the special estimation method (the product empirical distribution method) used by \cite{guo2021generalizing}? 
Specifically, we ask:
\begin{center}
\emph{If we use the standard empirical distribution method to estimate product distributions,\\
how many samples do we need?  Exponential or polynomial?}
\end{center} 

We show (in Section~\ref{sec:product_ERM_exp}) that the answer is ``exponential'' -- the empirical distribution method needs $\Omega(\frac{k^n}{\eps^2})$ samples, where $k$ is the size of support of each component distribution $D^j$ and $\eps$ is the desired estimation accuracy.  Thus, we conclude that, the exponential reduction of sample complexity from learning correlated distributions to learning product distributions relies on both the simpler structure of product distributions \emph{and} a specially designed estimation method that utilizes this structure.  Product distributions by themselves do not make learning easier.

\section{Distribution Learning}
In this section we define the problem of \emph{density estimation}, or \emph{distribution learning}, for discrete distributions. 
We focus on discrete distributions because density estimation for continuous distributions is complicated.  
Let $T=T^1\times \cdots \times T^n$ be an $n$-dimensional finite product space, with each component space $T^j$ having size $|T^j|=k$.  So, the size of $T$ is $\prod_{j=1}^n |T^j| = k^n$. 
Let $D$ be a distribution over $T$, so $D$ is a discrete distribution over $k^n$ points.
Given a set of $m$ i.i.d.~samples $S=\{s_1, \ldots, s_m\}$ from $D$, where each sample $s_i=(s_i^1, ..., s_i^n)\sim D$, our goal is to estimate $D$ by constructing, or learning, a distribution $E$ that is ``close'' to $D$ in a certain sense. 
In this note we measure the closeness between two distributions $E$ and $D$ using the total variation distance: 
\begin{equation}
    \dTV(E, D) = \frac{1}{2} \sum_{x\in T} | E(x) - D(x) |,
\end{equation}
where $D(x) = \Pr_{X\sim D}[X=x]$ denotes the probability assigned to $x$ by the distribution. 
The reason why we choose the total variation distance is because it characterizes the uniform convergence of all bounded functions $f: T\to [0, 1]$ in terms of the convergence of their means on $E$ to those on $D$: 
\begin{fact}\label{fact:d_TV}
$\dTV(E, D) \le \eps$ if and only if for all functions $f:T\to [0, 1]$, $| \E_{x\sim E} f(x) - \E_{x\sim D}f(x)| \le \eps$.
\end{fact}

We are interested in the \emph{sample complexity} of the distribution learning problem. That is, how many samples do we need to learn a distribution $E$ with $\dTV(E, D) \le \eps$ with high probability?
More formally, let $A: T^m \to \Delta(T)$ be a learning algorithm that takes $m$ samples $S\in T^m$ as input and outputs a distribution $E = A(S) \in \Delta(T)$, where $\Delta(T)$ denotes the set of all distributions on $T$. 
The \emph{sample complexity of algorithm $A$} is a function $M_A(\cdot, \cdot)$ of $\eps>0$ and $\delta\in(0, 1)$, such that, whenever the number of samples $m \ge M_A(\eps, \delta)$, we have
\begin{equation}
    \Pr_{S\sim D^m}[ \dTV(A(S), D) \le \eps] \ge 1-\delta.
\end{equation} 
The \emph{sample complexity of the distribution learning problem} is the minimum sample complexity over all learning algorithms: $M(\eps, \delta) = \inf_{A\,:\,\text{a learning algorithm}}{M_A(\eps, \delta)}$. 

\subsection{Exponential sample complexity of ERM to learn correlated distributions}\label{sec:correlated_exp}
When the distribution $D$ is an arbitrarily correlated distribution over the product space $T$, there is essentially no difference between $D$ and an arbitrary discrete distribution on $|T| = k^n$ points.  
It is known that the sample complexity of learning a discrete distribution on $K$ points is
$O\big(\frac{1}{\eps^2}(K + \log\frac{1}{\delta})\big)$
(see, e.g.,~\cite{Dia19Lecture_12}).
So, the sample complexity of learning arbitrarily correlated distributions is at most
\begin{equation}\label{eq:correlated_upper_bound}
    O\Big(\frac{1}{\eps^2}\big(k^n + \log\frac{1}{\delta}\big)\Big) = \tilde O\Big(\frac{k^n}{\eps^2}\Big) 
\end{equation}
(where the $\tilde O(\cdot)$ notation omits poly-logarithmic terms), which is exponential in the number of dimensions $n$. 
In particular, this sample complexity bound can be achieved by the simple and standard learning algorithm that outputs the \emph{empirical distribution}, which is the uniform distribution over samples:
\begin{equation}
  \emp(S) = \frac{1}{m} \sum_{i=1}^m \delta_{s_i},
\end{equation}
where $\delta_s$ denotes the distribution that puts probability $1$ on $s\in T$.  In order words, if $m \ge \tilde O(\frac{k^n}{\eps^2})$ then with probability at least $1-\delta$ we have $\dTV(\emp(S), D) \le \eps$. 

The above $\emp$ algorithm corresponds to the classical principle of Empirical Risk Minimization (ERM) in learning theory: in order to find a hypothesis $h^*$ in a hypothesis class $\mathcal H$ that minimizes the expected risk $\E_{x\sim D} h(x)$ with respect to the unknown distribution $D$, we can find the hypothesis $h^*_{\ERM}$ that minimizes the expected risk with respect to the empirical distribution $\emp(S)$, $\E_{x\sim \emp(S)} h(x) = \frac{1}{m} \sum_{i=1}^m h(s_i)$.  By Fact~\ref{fact:d_TV}, if $\dTV(\emp(S), D) \le \eps$, then $|\E_{x\sim \emp(S)} h(x) - \E_{x\sim D} h(x)| \le \eps$ for all hypotheses $h\in \mathcal H$ and hence it is easy to derive that $h^*_{\ERM}$ is a $2\eps$-approximation to $h^*$ in the sense that $\E_{x\sim D} h^*_{\ERM}(x) \le \E_{x\sim D}h^*(x) + 2\eps$. 

The exponential sample complexity upper bound in \eqref{eq:correlated_upper_bound} is tight in the sense that in order to learn any distribution $D$ over $T$ within total variation distance $\eps$, at least
\begin{equation}\label{eq:exp_lower_bound}
 \Omega\Big(\frac{1}{\eps^2}\big(k^n + \log\frac{1}{\delta}\big)\Big) = \tilde \Omega\Big(\frac{k^n}{\eps^2}\Big)
\end{equation}
samples are needed, no matter what the learning algorithm is.  This lower bound again follows from classical results on the sample complexity of learning discrete distributions (e.g., \cite{canonne_short_2020, Dia19Lecture}).

\subsection{Polynomial sample complexity of PERM to learn product distributions}
\label{sec:product_PERM_poly}

A surprising result by \cite{guo2021generalizing} shows that, when $D$ is a \emph{product} distribution rather than a correlated distribution, namely $D = D^1\times \cdots \times D^n$ where each $D^j$ is a distribution on $T^j$ and $D^j$'s are independent, the previous exponential sample complexity bound \eqref{eq:exp_lower_bound} of learning $D$ can be reduced to a polynomial: 
\begin{equation}
    m = O\Big(\frac{nk}{\eps^2} \log \frac{1}{\delta}\Big).
\end{equation}
However, the learning algorithm they use to prove this result is not the $\emp$ algorithm.  Instead, they use an algorithm that estimates $D$ by the following ``product empirical distribution'':
\begin{equation}
    \mathrm{Pemp}(S) = E^1 \times \cdots \times E^n,
\end{equation}
where each $E^j$ is the uniform distribution over the $m$ samples $\{s_1^j, \ldots, s_m^j\}$ from $D^j$.
They show that if the number of samples $m$ from each distribution $D^j$ is at least $O(\frac{nk}{\eps^2} \log \frac{1}{\delta})$, then with probability at least $1-\delta$, it holds that $\dTV(\mathrm{Pemp}(S), D) \le \eps$. 
Their $\mathrm{Pemp}$ algorithm gives rise to a new principle, called Product Empirical Risk Minimization (PERM), for general learning problems on product distributions: instead of using $h^*_{\ERM}$ to approximate the best hypothesis $h^*$ in a class $\mathcal H$, one should use $h^*_{\mathrm{PERM}}$ which is the best hypothesis with respect to the product empirical distribution: $h^*_{\mathrm{PERM}} = \argmin_{h\in\mathcal H} \E_{x\sim \mathrm{Pemp}(S)} h(x)$. 




\section{Exponential Sample Complexity of ERM to Learn Product Distributions}
\label{sec:product_ERM_exp}
Given the result by \cite{guo2021generalizing} that the (non-standard) $\mathrm{Pemp}$ algorithm is able to achieve a polynomial sample complexity for learning product distributions,
an immediate question is: does the standard $\emp$ algorithm achieve that as well? 
Namely, how many samples do we need for the empirical distribution to be $\eps$-close to $D$ in total variation distance?  In other words, if one insists on using ERM rather than PERM to learn a hypothesis class $\mathcal H$ on a product distribution, how is the generalization error of ERM compared to that of PERM? 

One may initially think that the aforementioned exponential sample complexity bound $\tilde \Omega(\frac{k^n}{\eps^2})$ of $\emp$ for learning arbitrarily correlated distributions \eqref{eq:exp_lower_bound} should be improved because product distributions have simpler structures than correlated distributions.  
However, that is not true as we show in the following theorem: $\emp$ still needs at least $\Omega(\frac{k^n}{\eps^2})$ samples to learn product distributions.  
\begin{theorem}\label{thm:ERM_exponential}
There exists a product distribution $D$ on the product space $T = T^1\times \cdots \times T^n$ with $|T^j|=k$, such that,
if
\[ m < c\frac{k^n}{\eps^2} \]
for some constant $0 < c < 0.017$, 
then with probability at least $1 - O(\frac{1}{k^n})$ over the random draw of $m$ i.i.d.~samples $S = \{s_1, \ldots, s_m\}$ from $D$, we have $\dTV(\emp(S), D) > \eps$. 
\end{theorem}

This theorem demonstrates a fundamental difference between the $\emp$ algorithm and the $\mathrm{Pemp}$ algorithm: the former ignores the structure of product distributions; the latter makes use of the independence of different dimensions of the distribution $D$ by first estimating each dimension of $D$ independently and then taking the product.  This difference leads to an exponential gap in their sample complexities. \\

\noindent \textbf{Remark 1.}  
The product distribution $D$ we use in the proof of Theorem~\ref{thm:ERM_exponential} is simple: it is the uniform distribution over all the $k^n$ points in $T$.  \\

\noindent \textbf{Remark 2.}
The following ``expectation'' version of Theorem~\ref{thm:ERM_exponential} is much easier to prove: if $m < c\frac{k^n}{\eps^2}$, then $\E_{S\sim D^m}[\dTV(\emp(S), D)] \ge \Omega(\eps)$.  The ``with high probability'' version we presented above is stronger and more difficult to prove. \\  

Before proving Theorem~\ref{thm:ERM_exponential} formally in Section~\ref{sec:proof}, we provide a simple and intuitive argument to show a weaker bound of $\Omega(k^n)$ rather than $\Omega(\frac{k^n}{\eps^2})$.   
Let $D$ be the uniform distribution over $T$, so $D(x) = \frac{1}{k^n}$ for any $x\in T$.  Suppose we have $m\ll k^n$ samples $S=\{s_1, \ldots, s_m\}$ from $D$.  Because $m$ is much smaller than $k^n$ the samples in $S$ are with high probability distinct. 
Let $E = \emp(S)$ be the empirical distribution, so it puts $\frac{1}{m}$ probability on each point $x$ in $S$ and $0$ probability outside of $S$.  Then, the total variation distance between $E$ and $D$ is
$\dTV(E, D) = \frac{1}{2} \sum_{x\in T} |E(x) - D(x)| \ge \frac{1}{2} \sum_{x\in S} (\frac{1}{m} - \frac{1}{k^n}) = \frac{1}{2}(1 - \frac{m}{k^n})$. This is much larger than $\eps$ given $m\ll k^n$. 


\subsection{Proof of Theorem~\ref{thm:ERM_exponential}}
\label{sec:proof}
The main idea of the proof is the \emph{Poisson approximation} in the ``balls and bins'' problem in the literature of randomized algorithms \cite{mitzenmacher2001power}.
Let $D$ be the uniform distribution over $T$. 
To simplify notations we let $N = |T| = k^n$ and regard $D$ as the uniform distribution over $N$ discrete points $\{1, 2, \ldots, N\}$.  Let $X_1, \ldots, X_N$ be $N$ random variables where each $X_j=\sum_{i=1}^m \mathbbm 1[s_i = j]$ records the number of samples that are equal to point $j$ among the total $m$ samples.
Notice that $X_1, \ldots, X_N$ are correlated and their sum is $\sum_{j=1}^N X_j = m$.
Marginally, each $X_j$ follows the binomial distribution $\mathrm{Bin}(m, \frac{1}{N})$ with expectation $\E[X_j] = \frac{m}{N}$ and is related to the empirical distribution $E = \emp(S)$ by the equation $X_j = m E(j) = \sum_{i=1}^m \mathbbm 1[s_i = j]$.
Thus, the event $\mathcal{E} = \mathbbm 1[ \dTV(E, D) < \eps]$ 
can be equivalently expressed as
\begin{align*}
  \mathbbm 1[ \dTV(E, D) < \eps] ~\iff~ \mathbbm 1\Big[ \sum_{j=1}^N |E(j) - \tfrac{1}{N}| < 2\eps\Big] ~ \iff~ \mathbbm 1\Big[ \sum_{j=1}^N |X_j - \tfrac{m}{N}| < 2m\eps\Big].
\end{align*}
We want to show that, if $m < c \frac{N}{\eps^2}$, then 
\begin{equation}\label{eq:dtv_event_goal}
    \Pr[ \dTV(E, D) < \eps ] = \Pr\Big[\sum_{j=1}^N |X_j - \tfrac{m}{N}| < 2m\eps\Big] < O\Big(\frac{1}{N}\Big). 
\end{equation}
The challenging part in analyzing $\Pr\big[\sum_{j=1}^N |X_j - \tfrac{m}{N}| < 2m\eps\big]$ is that the random variables $X_j$'s are not independent.  The idea of Poisson approximation is to approximate $X_1, \ldots, X_N$ by \emph{independent} Poisson random variables $Y_1, \ldots, Y_N$, where each $Y_j$ follows $\mathrm{Poisson}(\lambda)$ distribution with parameter $\lambda = \frac{m}{N}$, so that $Y_j$ has the same expected value as $X_j$: $\E[Y_j] = \lambda =  \frac{m}{N} = \E[X_j]$.
The approximation is formalized by the following lemma: 
\begin{lemma}[Poisson approximation, Theorem 2.11 in \cite{mitzenmacher2001power}]\label{lem:poisson}
Let $\mathcal E(X_1, \ldots, X_N)$ be an event of the $N$ random variables $X_1, \ldots, X_N$ such that the probability $\Pr_X\big[\mathcal E(X_1, \ldots, X_N)\big]$ is monotonically increasing or decreasing in $m=\sum_{j=1}^N X_j$.  Then, $\Pr_X[ \mathcal E(X_1, \ldots, X_N) ] \le 4 \Pr_Y[ \mathcal E(Y_1, \ldots, Y_N)]$.
\end{lemma}
We use Lemma~\ref{lem:poisson} for the event in \eqref{eq:dtv_event_goal}, $\mathcal E = \mathcal E(X_1, \ldots, X_N) = \mathbbm 1[\sum_{j=1}^N |X_j - \tfrac{m}{N}| < 2m\eps]$.  It is not hard to show that the probability $\Pr_X[\mathcal E(X_1, \ldots, X_N)] = \Pr_X[\sum_{j=1}^N |X_j - \tfrac{m}{N}| < 2m\eps]$ is monotonically increasing in $m$.\footnote{The intuition is that, as the number of samples $m$ increases, the random variable $\frac{X_j}{m}$ is more concentrated around its mean $\frac{1}{N}$, so the equivalent event $\sum_{j=1}^N |\frac{X_j}{m} - \tfrac{1}{N}| < 2\eps$ is more likely to happen.}
So, we can use Lemma~\ref{lem:poisson} to derive that 
\begin{equation}
    \Pr_X\Big[\sum_{j=1}^N |X_j - \tfrac{m}{N}| < 2m\eps\Big] \le 4 \Pr_Y\Big[\sum_{j=1}^N |Y_j - \tfrac{m}{N}| < 2m\eps\Big]. 
\end{equation}
It remains to prove $\Pr_Y[\sum_{j=1}^N |Y_j - \tfrac{m}{N}| < 2m\eps] \le O(\frac{1}{N})$.  We use Chebyshev's inequality to do so.  We will use the following lemma, whose proof is in Appendix~\ref{app:a}.
\begin{lemma}\label{lem:Y}
Suppose $\lambda=\frac{m}{N} \ge 2$.\footnote{When $\lambda < 2$, i.e., $m < 2N = 2k^n$, one can prove Theorem~\ref{thm:ERM_exponential} using a similar argument to the simple argument given before the formal proof.}
The expectation and variance of $|Y_j-\frac{m}{N}|$, where $Y_j\sim \mathrm{Poisson}(\lambda)$, are bounded by: (1) $\E[|Y_j - \frac{m}{N}|] \ge \frac{1}{\sqrt \pi e^{1/12}}\sqrt{\lambda}$; (2) $\Var(|Y_j - \frac{m}{N}|) \le \lambda$. 
\end{lemma}
Let $c_1 = \frac{1}{\sqrt \pi e^{1/12}}$, so $\E[|Y_j - \frac{m}{N}|] \ge c_1\sqrt{\lambda}$. 
Since $Y_1, \ldots, Y_N$ are independent, we have
\begin{align}
    \E\Big[\sum_{j=1}^N |Y_j - \tfrac{m}{N}|\Big] \ge N c_1\sqrt{\lambda} = c_1\sqrt{mN} ~~~ \text{and}~~~
    \Var\Big(\sum_{j=1}^N |Y_j - \tfrac{m}{N}|\Big) \le N \lambda = m. 
\end{align}
It is easy to verify that, when $m < \frac{c_1^2}{4\eps^2} N$, we have $2m\eps < c_1\sqrt{mN} \le \E[\sum_{j=1}^N |Y_j - \tfrac{m}{N}|]$ and hence by Chebyshev's inequality, 
\begin{align}\label{eq:prob}
    \Pr\bigg[\,\sum_{j=1}^N |Y_j - \tfrac{m}{N}| < 2m\eps\,\bigg] & \le \Pr\Bigg[ \, \bigg|\sum_{j=1}^N |Y_j - \tfrac{m}{N}| - \E\Big[\sum_{j=1}^N |Y_j - \tfrac{m}{N}|\Big] \bigg| > c_1\sqrt{mN} - 2m\eps \,\Bigg] \nonumber \\
    & \le \frac{m}{(c_1\sqrt{mN} - 2m\eps)^2 }. 
\end{align}
Suppose $m = c \frac{N}{\eps^2}$ with $c < \frac{c_1^2}{16}\approx 0.017$.  Plugging into \eqref{eq:prob} finishes the proof:  
\begin{align*}
     \frac{m}{(c_1\sqrt{mN} - 2m\eps)^2 } = \frac{\tfrac{cN}{\eps^2}}{(c_1\sqrt c \tfrac{N}{\eps} - 2c \frac{N}{\eps})^2} = \frac{c}{(c_1\sqrt c - 2c)^2}\frac{1}{N} \le \frac{4}{c_1^2}\frac{1}{N}  = O\Big(\frac{1}{N}\Big). 
\end{align*}


\section{Conclusion and Future Directions}
We conclude that, although (discrete) product distributions are simpler than correlated distributions, they cannot be learned efficiently unless we use algorithms that are specifically designed for them (e.g., PERM). 
We conjecture that similar conclusions also hold in the distribution learning problem for continuous distributions.  

The distribution learning problem we consider here is independent of the hypothesis class.
An interesting future direction could be to combine complexity measures of hypothesis classes (e.g., VC dimension) and the product structure of a distribution to show a hypothesis-dependent sample complexity bound for product distributions that is smaller than the classical hypothesis-dependent distribution-free bounds.

%% file: appendix.tex
\appendix
\section*{Appendix}

\section{Proof of Lemma~\ref{lem:Y}} \label{app:a}
Let $Y_j\sim \mathrm{Poisson}(\lambda=\frac{m}{N})$.  We first show the (2) term $\Var(|Y_j - \tfrac{m}{N}|) \le \lambda$: 
\begin{align*}
    \Var(|Y_j - \tfrac{m}{N}|) & = \E[|Y_j - \tfrac{m}{N}|^2] - (\E[|Y_j - \tfrac{m}{N}|])^2 \\
    & \le  \E[|Y_j - \tfrac{m}{N}|^2] \\
    & = \Var(Y_j) = \lambda. 
\end{align*}
Then we consider the (1) term $\E[|Y_j - \tfrac{m}{N}|]$.  According to the property of Poisson distribution, we have $\E[|Y_j - \tfrac{m}{N}|] = \frac{2\lambda^{\lfloor \lambda\rfloor + 1}e^{-\lambda}}{\lfloor \lambda \rfloor!}$.\footnote{See, e.g., \url{https://en.wikipedia.org/wiki/Poisson_distribution}}
So, by Stirling's approximation $n! \le \sqrt{2\pi n}(\tfrac{n}{e})^n e^{1/12n}$ and the fact that $2x^{\lfloor \lambda \rfloor + 1}e^{-x}$ is increasing in $x$ when $x\le \lambda$, we obtain  
\begin{align*}
   \E[|Y_j - \tfrac{m}{N}|] & \ge \frac{2\lfloor\lambda\rfloor^{\lfloor \lambda\rfloor + 1}e^{-\lfloor\lambda\rfloor}}{\lfloor \lambda \rfloor!} \\
   & \ge \frac{2\lfloor\lambda\rfloor^{\lfloor \lambda\rfloor + 1}e^{-\lfloor\lambda\rfloor}}{\sqrt{2\pi \lfloor \lambda \rfloor} (\tfrac{\lfloor \lambda \rfloor}{e})^{\lfloor \lambda \rfloor}e^{1/(12\lfloor \lambda \rfloor)}} \\
   & = \sqrt{\frac{2}{\pi}} \frac{\sqrt{\lfloor \lambda \rfloor}}{e^{1/(12\lfloor \lambda \rfloor)}} \\
   & \ge \sqrt{\frac{2}{\pi}} \frac{\sqrt{ \lambda / 2}}{e^{1/12}} \\
   & = \frac{1}{\sqrt\pi e^{1/12}} \sqrt{\lambda}.  
\end{align*}